\documentclass{Interspeech}

\interspeechcameraready 

\title{DiceHuBERT: Distilling HuBERT with a Self-Supervised Learning Objective}

\author[affiliation={1}]{Hyung Gun}{Chi}
\author[affiliation={1}]{Zakaria}{Aldeneh}
\author[affiliation={1}]{Tatiana}{Likhomanenko}
\author[affiliation={1}]{Oggi}{Rudovic}
\author[affiliation={1}]{Takuya}{Higuchi}
\author[affiliation={2}]{Li-Wei}{Chen}
\author[affiliation={2}]{Shinji}{Watanabe}
\author[affiliation={3, \dagger}]{Ahmed Hussen}{Abdelaziz}

\affiliation{Apple}{$^2$Carnegie Mellon University}{$^3$Meta}
\email{hchi23@apple.com}

\usepackage{comment}
\usepackage{multirow}
\usepackage[table,xcdraw]{xcolor}
\usepackage{subcaption}
\usepackage{hyperref}
\usepackage{cite}
\usepackage{url}
\usepackage[capitalize]{cleveref}

\begin{document}

\maketitle

\renewcommand{\thefootnote}{}
\footnotetext{$^\dagger$Work done while at Apple.}

\begin{abstract}
We introduce DiceHuBERT, a knowledge distillation framework for compressing HuBERT, a widely used self-supervised learning (SSL)-based speech foundation model. Unlike existing distillation methods that rely on layer-wise and feature-wise mapping between teacher and student models, DiceHuBERT leverages  HuBERT's iterative self-distillation mechanism by directly replacing the original model with a student model. This replacement allows the student to be trained using the same SSL objective used when pre-training HuBERT, eliminating the need for additional modules or architectural constraints. Experimental results on SUPERB show that DiceHuBERT consistently outperforms existing distillation methods, improving phoneme recognition performance by over $21\%$ and ASR performance by more than $14\%$. Furthermore, DiceHuBERT demonstrates competitive performance across multiple tasks, highlighting its clear advantage.
\end{abstract}

\section{Introduction}

Self-supervised learning (SSL)-based speech foundation models \cite{baevski2020wav2vec, hsu2021hubert, chen2022wavlm, babu2021xls, song2020speech, huang2022investigating, chang2021exploration, chung2021w2v} have received significant attention due to their ability to generate general-purpose speech representations for a wide range of speech tasks. However, the substantial size of the SSL-based foundation models makes them unsuitable for deployment in mobile applications \cite{sainath2015convolutional, wu2018monophone}. Knowledge distillation \cite{hinton2015distilling, gou2021knowledge, ba2014deep} is a widely adopted technique in various domains \cite{chen2017learning, cui2017knowledge, yang2020distilling, asami2017domain, gandhi2023distil, chen2022knowledge} that effectively compresses the model size while maintaining performance. This compression is achieved by transferring knowledge from a larger ``teacher'' model to a smaller ``student'' model.

Recent efforts \cite{peng2023dphubert, lee2022fithubert, jang2023recycle, chang2022distilhubert, jang2024star, wang2022lighthubert} in knowledge distillation for SSL-based speech foundation models, particularly HuBERT\cite{hsu2021hubert}, have focused on compressing architectures and introducing new loss functions. These methods typically distill knowledge by aligning layer-wise features between the teacher and student models. However, they often overlook the original SSL objective and impose architectural constraints on the student model. 

\begin{figure*}[t]
    \centering
    \begin{subfigure}[b]{0.45\linewidth}
        \centering
        \includegraphics[width=\linewidth]{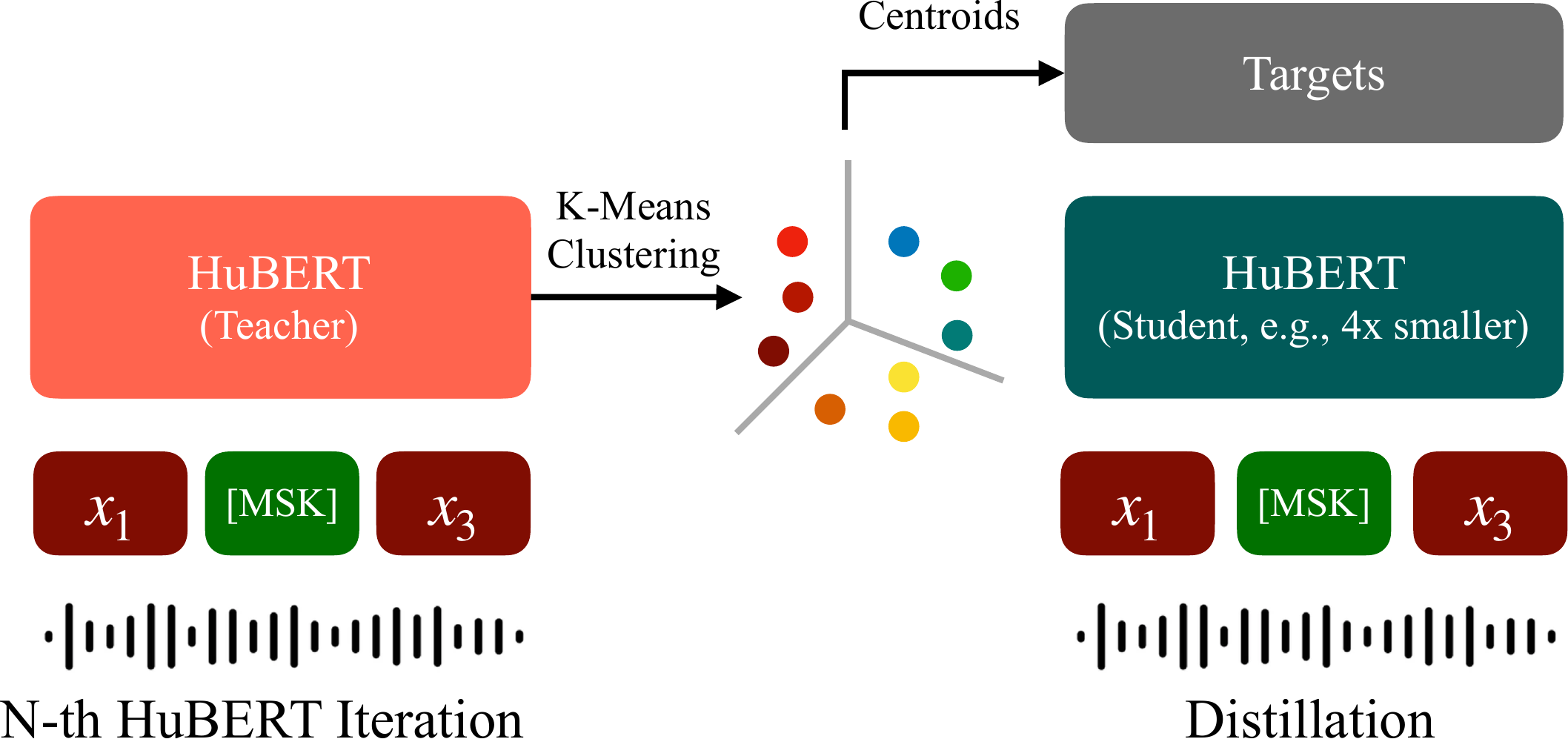}
        \caption{DiceHuBERT}
    \end{subfigure}
    \hfill
    \begin{subfigure}[b]{0.45\linewidth}
        \centering
        \includegraphics[width=\linewidth]{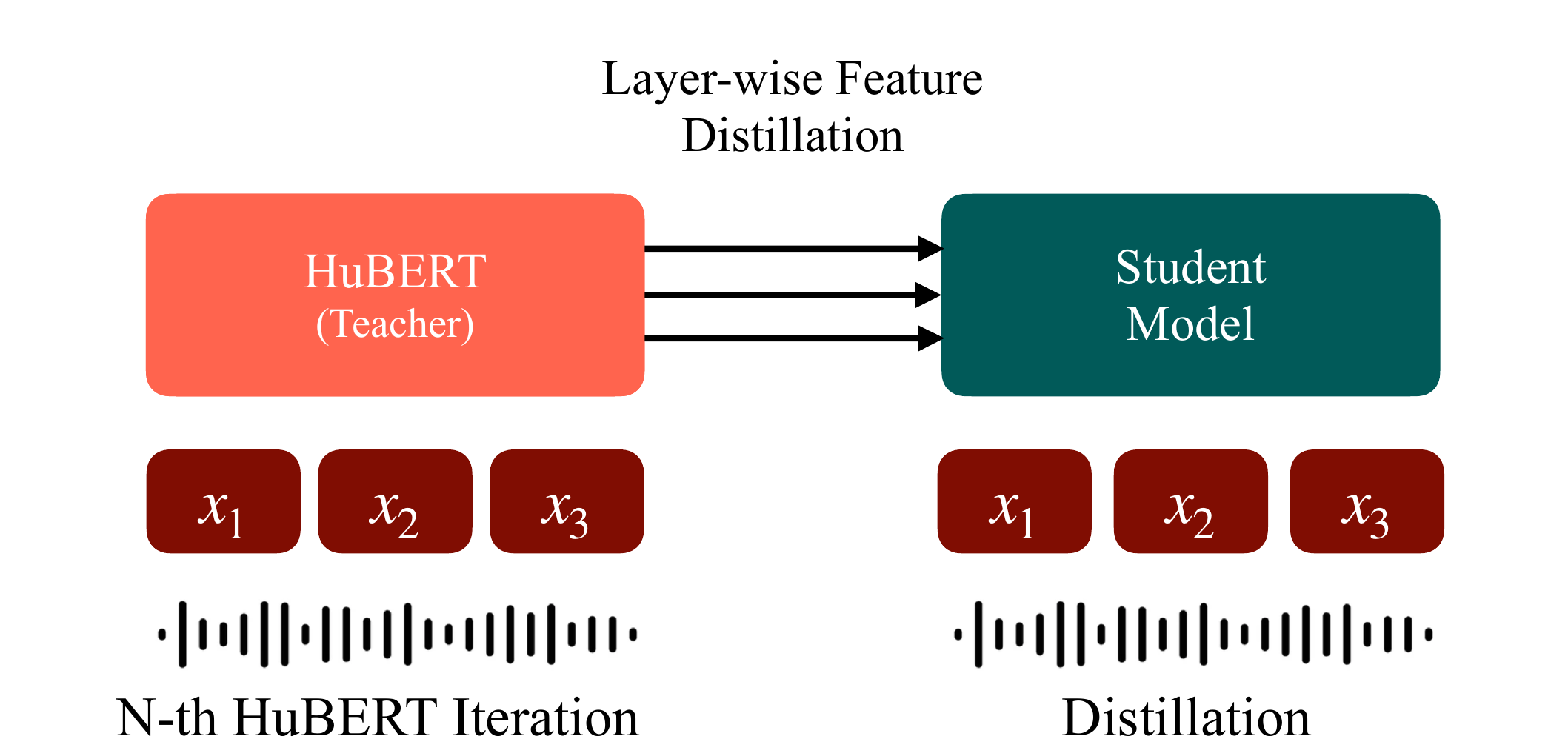}
        \caption{Feature Distillation}
    \end{subfigure}
    \vspace{-0.5em}
    \caption{DiceHuBERT differs from feature distillation-based methods in that it distills knowledge from target labels generated by a teacher model using $k$-means clustering. This process allows DiceHuBERT to be trained using an SSL objective that closely resembles the training objective of the teacher model.}
    \label{fig:main}
    \vspace{-1em}
\end{figure*}

In this paper, we introduce DiceHuBERT, \textbf{Di}stilling from cluster \textbf{Ce}ntroids HuBERT, to address the aforementioned limitations. DiceHuBERT is trained using the same SSL objective as the teacher model, with knowledge distillation occurring through targets generated from the $k$-means \cite{arthur2006k} centroids of the teacher model's features, as illustrated in \Cref{fig:main} (a). DiceHuBERT is a simple yet effective distillation framework. Unlike existing methods that primarily focus on feature distillation, see \Cref{fig:main} (b), DiceHuBERT leverages the iterative self-distillation of HuBERT's pre-training, where knowledge is progressively distilled from each iteration to the next. Building upon this inherent self-distillation mechanism, \textit{DiceHuBERT is a natural extension of HuBERT that enables efficient compression within the HuBERT framework itself, without the need for additional modules.} 

In our approach, the student model architecture is unrestricted because distillation relies solely on the target labels produced by the teacher model. In contrast, feature distillation methods typically require layer-wise or feature-wise mapping between teacher and student representations, which imposes design constraints on the student model. This architectural flexibility, combined with the reuse of the original HuBERT self-distillation framework, makes DiceHuBERT more straightforward, practical, and easier to adopt than previous methods.

We compare our SSL-based distillation with feature distillation methods from existing works on four tasks from SUPERB \cite{yang2021superb}: phoneme recognition (PR), automatic speech recognition (ASR), speaker identification (SID), and automatic speaker verification (ASV). Our results show that SSL-based distillation consistently outperforms feature distillation across various SUPERB downstream tasks. Specifically, our proposed distillation approach outperforms prior HuBERT distillation methods, achieving over 21\% relative improvement in PR and more than 14\% relative improvement in ASR. Improvements are also observed in ASV and SID. We further investigate different architectures and label choices for the SSL objective during distillation, offering insights to inform future research in this domain.




 \label{sec:intro}
\section{Preliminaries}

\subsection{Prior HuBERT Distillation} \label{sec:feat}
Most prior works on HuBERT distillation \cite{gandhi2023distil, lee2022fithubert, jang2024star, jang2023recycle} focus on transferring knowledge from the pre-trained second-iteration HuBERT model (used as the teacher) to the student model through layer-wise feature distillation, as shown in \Cref{fig:main} (b)\footnote{Note that \Cref{fig:main} (b) illustrates general layer-wise feature distillation, which differs among models. e.g., MaskHuBERT \cite{jang2023recycle} employs masked inputs, DistilHuBERT \cite{chang2022distilhubert} incorporates prediction heads.}.
The student model usually has the same number of layers as the teacher, enabling a direct correspondence between their layer outputs for distillation, while reducing the feature dimensions. 
The distillation objective is commonly defined as the mean squared error (MSE) between the teacher’s and student’s features. 
To align the student ($S$) feature dimension $D^S$ with the teacher ($T$) feature dimension $D^T$, a linear projection layer $W_{(l)} \in \mathbb{R}^{D^S \times D^T}$ is employed:
\begin{align}
&\mathcal{L}_{\text{feat}}  = \sum_{l\in\{1, \ldots, L\}} \alpha_{(l)} \text{MSE}(H_{(l)}^T, H_{(l)}^S W_{(l)}), \label{eq:distil}
\end{align}
where  $H_{(l)}$ is a feature from $l$-th layer and $\alpha_{(l)}$ is weight for each layer loss. 

Although student model architectures vary, and some works add additional loss functions to enhance alignment, previous work generally follows the above objective for HuBERT distillation.

\subsection{HuBERT Pre-training}
HuBERT \cite{hsu2021hubert} is an SSL speech representation approach that employs a masked token prediction task. The pre-training process involves $N$ iterations, each with distinct targets but the same learning objective predicting the targets of input features corresponding to randomly masked timestamps in the audio. 
Formally, the SSL loss is defined as the cross-entropy loss computed over the masked timestamps:
\begin{align}
\mathcal{L}_{\text{SSL}} = -\sum_{t \in M} \log p(z_t|\tilde{X}, t), \label{eq:ssl}
\end{align}
where $M$ denotes the set of masked timestamps, $\tilde{X}$ is a masked input from a speech utterance $X\!=\![x_1, \ldots, x_T]$ using $M$, and $z_t$ is a target class at timestamp $t$.
The targets are generated through $k$-means clustering: in the first iteration, MFCC features are used; in the following iteration $N$, typically the $6$th layer hidden features from the previous iteration $(N\!-\!1)$th model serve as the input for clustering to generate new targets for an  $N$th model. The each following iteration can be viewed as a \textit{self-distillation} process as it uses the pre-trained features from the previous iteration to guide the model's training, while keeping the model architecture the same with the first iteration.
 \label{sec:background}

\section{DiceHuBERT}
We propose DiceHuBERT, a HuBERT distillation framework that offers a simpler and natural distillation approach within HuBERT itself compared to previous methods.  DiceHuBERT leverages HuBERT’s iterative self-distillation process by reducing the model size when transitioning from iteration $N$ (teacher) to iteration $N\!+\!1$ (student), all while maintaining the original SSL objective, see \Cref{fig:main} (a).
DiceHuBERT offers two main advantages over existing HuBERT distillation methods. Firstly, it eliminates the need for an additional distillation framework, allowing us to directly reuse the pre-training framework of HuBERT without implementing any new modules. As a result, DiceHuBERT is architecture-agnostic, providing flexibility in choosing the student model architecture. The feature distillation approach's reliance on layer-wise feature matching hinders flexibility in student model design. Secondly, our approach utilizes the same SSL objective as the teacher model, which has been shown to be more effective than MSE loss from \cref{eq:distil} for learning robust representations \cite{menon2020distillation}.  In the following sections, we focus on two key aspects that can impact distillation performance of DiceHuBERT: the student architecture and the targets used for the SSL objective.

\subsection{Student Architecture} \label{sec:archi}
We propose student architectures to examine two key aspects of DiceHuBERT: 1) the feature dimension $D^S$ and 2) the number of layers $L$. To isolate the effects of these factors, we design our student model architecture, keeping all other elements unchanged. 
We build two types of student models, HuBERT-shallow and HuBERT-narrow, by modifying the original HuBERT-base model. HuBERT-shallow is obtained by reducing \textit{\textbf{only} the number of layers}.
HuBERT-narrow, on the other hand, reduces \textit{\textbf{only} the feature dimension} and \textit{the intermediate feed-forward dimension}.
Note that DiceHuBERT can be used with any model, including convolutional-only designs.

\subsection{Target Labels} \label{sec:label}
We define two distinct target labels for our SSL objective. The first is hard labels, generated through $k$-means clustering of intermediate features from the HuBERT model of $N$-th iteration. These labels are one-hot encoded, and we can directly apply cross-entropy loss following \cref{eq:ssl}.
The second target is soft labels, which provide richer information than hard labels and enhance knowledge distillation \cite{hinton2015distilling}. Soft labels are obtained by calculating the distance of each feature to all centroids generated by $k$-means and further applying softmax to estimate each cluster probability, where hard labels are computed via argmin of these distances. The probability $p(i|H)$ of $i$-th cluster with centroid $H_i$ for the feature $H$ is defined as:
\begin{align}
p(i|H) = \frac{\exp(-\rho(H, H_i )/\tau)}{\sum_{j=\{1,..,K\}} \exp(-\rho(H, H_j)/\tau)},
\label{eq:softlabel}
\end{align}
where $\rho(\cdot, \cdot)$ is the $L_2$ distance between two representations, $K$ is the number of clusters, and $\tau$ is the temperature parameter. 
Then, the SSL loss for soft labels is defined as the KL-divergence between the output class distribution of a student model and the soft labels from a teacher model. \label{sec:method}
\section{Experiments}
\subsection{Implementation Details} \label{sec:impl}
Since there are no publicly available pre-train weights for the first iteration of HuBERT, we train this model from scratch using the hyper-parameters reported in the HuBERT paper~\cite{hsu2021hubert}.
For our distillation process, we follow exactly the same setup of HuBERT second iteration pre-training, while changing only the model architecture to student following Section~\ref{sec:archi}. 
For example, we set the mask span to 10 and randomly select 8\% of the waveform encoder output frames. 
As training data, we use LibriSpeech 960h \cite{panayotov2015librispeech} following prior works. All our experiments are conducted on 32 NVIDIA A100 40GB GPUs.  We used the same setup for all our experiments, unless otherwise stated.

\subsection{Evaluation} 
We evaluate DiceHuBERT on SUPERB~\cite{yang2021superb}, focusing our evaluation on four tasks that exhibit the significant performance gaps between the teacher and student models: content-based tasks (PR, ASR) and speaker-based tasks (SID, ASV).


\subsection{Comparison with State-of-the-Art Methods}
We compare DiceHuBERT with other state-of-the-art HuBERT distillation methods in \Cref{table:main}. For this experiment, we use the HuBERT-narrow ($\!D_{\text{base}}/2$) setup as the student architecture, and train our model solely with the SSL loss. For evaluation, we use the default SUPERB recipes but modify the learning rates for the SID task to 5e-3 following prior works \cite{peng2023dphubert, jang2024star, lee2022fithubert} for a fair comparison.
All distillation methods in \Cref{table:main}, including our method, use the second iteration of HuBERT-base as the teacher model.
DiceHuBERT outperforms all other distillation methods across all tasks. Notably, DiceHuBERT achieves significant improvements in content-based tasks, such as a 21\% relative improvement in ASR and a 14\% relative improvement in PR, compared to the previous best models. In contrast, the performance gap in speaker-based tasks is not big but surpasses the other approaches, with improvements of 3.6\% in ASV and 0.7\% in SID. This comparison demonstrates the advantages of DiceHuBERT over feature distillation-based methods.

\begin{table}[t]
\centering
\caption{Comparison with SOTA HuBERT distillation methods. All distillation methods, including our proposed method, use the second iteration of HuBERT-base as the teacher. We exclude LightHuBERT \cite{wang2022lighthubert} as its compression process requires extensive computational cost.}
\resizebox{1.0\linewidth}{!}{
\addtolength{\tabcolsep}{-4.5pt}
\begin{tabular}{lccccc}
\toprule
\multicolumn{1}{c}{} & \# Param & PR & ASR (w/o LM) & ASV & SID  \\
\multicolumn{1}{c}{\multirow{-2}{*}{Methods}} & M ↓ & PER ↓  & WER ↓ & EER ↓ & ACC ↑ \\\hline\hline 

\multicolumn{6}{c}{\textit{Baselines}}\\
FBank & - & 82.01 & 23.18 & 9.56 &8.5e-4 \\
Wav2Vec2-base \cite{baevski2020wav2vec} & 95 & 5.74 & 6.43 & 6.02 & 75.18 \\
HuBERT-base \cite{hsu2021hubert}  & 94 & 5.41 & 6.42 & 5.11 & 81.42 \\

\midrule
\multicolumn{6}{c}{\textit{Distillation Methods}} \\
DistilHuBERT \cite{chang2022distilhubert} & 23 & 16.27 & 13.37 & 8.55 & 73.54 \\
FitHuBERT \cite{lee2022fithubert} & 22 & 13.32 & 12.09 & 8.00 & 55.71  \\
DPHuBERT \cite{peng2023dphubert} & 23 & 9.67 & 10.47 & 5.84 & 76.83  \\
MaskHuBERT \cite{jang2023recycle} & 26 & 7.30 & 9.77 & 5.38 & 62.83 \\
ArmHuBERT \cite{jang2023recycle} & 26 & 7.72 & 9.96 & 5.68 & 65.03 \\
StarHuBERT-L \cite{jang2024star} & 26 & 7.97 & 8.91 & 5.45 & 78.66 \\ 
\midrule
\multicolumn{6}{c}{\textit{Proposed Method}}\\
\textbf{DiceHuBERT} & 26 & \textbf{6.23} & \textbf{7.64}& \textbf{5.25} & \textbf{79.23} \\
\bottomrule
\end{tabular}
}
\label{table:main}
\vspace{-0.4cm}
\end{table}

\begin{table}[t]
\centering
\caption{Performance comparison across different student model architectures. $L_{\text{base}}$ and $D_{\text{base}}$ represent the number of layers and feature dimension of the HuBERT-base, respectively.}
\resizebox{1.0\linewidth}{!}{
\addtolength{\tabcolsep}{-3pt}
\begin{tabular}{lccccc}
\toprule
& \#Param & PR    & ASR (w/o LM) & ASV & SID \\
\multirow{-2}{*}{Models} & M ↓ & PER ↓ & WER ↓ / UER ↓ & EER ↓ & ACC ↑\\ \hline\hline 
\multicolumn{6}{c}{\textit{Baselines}}\\
Upper Bound & 94 & 5.41 & 6.42 / -\hspace{1.5em}   & 5.11 & 81.42 \\
Lower Bound & 26 & 7.91 & 8.44 / 2.46 & 5.32 & 77.40\\ 
\midrule
\multicolumn{6}{c}{\textit{HuBERT-Shallow}}\\
$L \!=\! L_{\text{base}}/6$         & 23 & 14.45 & 13.17 / 3.99 &7.37  & 65.78\\
$L \!=\! L_{\text{base}}/4$         & 30 & 10.95 & 11.80 / 3.55 &6.55  & 72.10\\
\midrule
\multicolumn{6}{c}{HuBERT-Narrow}\\
$D^S \!=\! D_{\text{base}}/4$          & 9  & 12.55 & 11.17 / 3.36 & 6.24 & 57.64\\
$D^S \!=\! D_{\text{base}}/2$          & 26 &  \textbf{7.48} & \textbf{7.97} / \textbf{2.35} &\textbf{5.03}  & \textbf{76.85}\\
\bottomrule
\end{tabular}
}
\label{table:model_archi}
\vspace{-0.4cm}
\end{table}

\subsection{Ablation Studies}
We conduct ablation studies to investigate the impact of different configurations of DiceHuBERT on distillation.  We begin by exploring two approaches to reducing model size: decreasing the feature dimension and reducing the number of layers. We then compare our proposed SSL-based distillation with feature distillation while fixing the architecture. Next, we compare different targets for the distillation loss: soft labels and hard labels. Lastly, we investigate the impact of teacher model size by using targets from different teacher sources.
Unless otherwise specified, we use a narrow model ($D_{\text{base}}/2$) as the student architecture, HuBERT first-iteration model ($N\!=\!1$) as the teacher, employing hard labels as our default setup for ablation studies. 

\vspace{.5em}\noindent\textbf{Baselines.} \label{sec:baseline}
We establish performance upper and lower bound for our ablation studies in \Cref{table:model_archi}. To ensure a fair comparison, we select models of second iteration ($N\!=\!2$) for our baselines.
The upper bound is defined by the performance of the teacher model, which is the second iteration of the HuBERT-base model. To establish a lower bound that is \textbf{absent from all prior works}, we train HuBERT-narrow ($\!D_{\text{base}}/2$) from scratch following HuBERT-base pre-training process with two iterations. All the hyperparameters are identical with HuBERT-base pretraining except for the model architecture.  Surprisingly, our lower bound outperforms all other distillation methods in \Cref{table:main} in terms of ASR and ASV, while achieving comparable results on the other tasks. Previous works have primarily focused on comparing the distilled model performance with that of teacher model, omitting this essential lower bound. The lower bound is vital because it illuminates the extent of improvement achieved through distillation compared to training the student model from scratch. 

\vspace{.5em}\noindent\textbf{Student Architectures.}
In \Cref{table:model_archi}, we compare distillation performance of our framework on different student architecture that we described in \cref{sec:archi}. For simplicity, `Shallow' and `Narrow' represent HuBERT-shallow and HuBERT-narrow, respectively. We first compare shallow ($\!L_{\text{base}}/6$) and narrow ($\!D_{\text{base}}/4$). Narrow ($\!D_{\text{base}}/4$) outperforms shallow ($\!L_{\text{base}}/6$) on PR, ASR, and ASV, even though narrow ($\!D_{\text{base}}/4$) is 2.5 times smaller than shallow ($\!L_{\text{base}}/6$). Next, we compare shallow ($\!L_{\text{base}}/4$) and narrow ($\!D_{\text{base}}/2$). Remarkably, narrow ($\!D_{\text{base}}/2$) demonstrates significantly better performance than shallow ($\!L_{\text{base}}/4$), despite the fact that shallow ($\!D_{\text{base}}/2$) has a smaller model size. These results shows importance of maintaining the number of layers for model compression. In SUPERB, it incorporates a weighted sum of the intermediate features of the models for downstream tasks. Reducing the number of layers would result in the loss of layer-wise knowledge distilled from the teacher, leading to poor performance for SUPERB. This finding partially explains the inferior performance of DistilHuBERT \cite{gandhi2023distil}, which shares an identical architecture with a shallow ($\!D_{\text{base}}/6$) model.


\begin{table}[t]
\centering
\caption{Performance comparison across different distillation methods.}
\vspace{-0.5em}
\addtolength{\tabcolsep}{-3pt}
\begin{tabular}{lcccc}
\toprule
& PR    & ASR (w/o LM) & ASV  & SID \\
\multirow{-2}{*}{Losses} & PER ↓ & WER ↓ / UER ↓        & EER ↓ & ACC ↑\\ \hline\hline 
$\mathcal{L}_\text{SSL}$ & \textbf{7.48}  & \textbf{7.97} / \textbf{2.35} & \textbf{5.03}& 76.85\\
$\mathcal{L}_\text{SSL} + 0.1 \cdot \mathcal{L}_\text{feat}$ &7.64  & 8.28 / 2.39 &5.36  & \textbf{78.01}\\
$\mathcal{L}_\text{SSL} + 1 \cdot \mathcal{L}_\text{feat}$   &8.97  & 8.71 / 2.52 &5.28  & 76.98\\
$\mathcal{L}_\text{feat}$ & 8.19 & 9.97 / 2.95 & 5.24 & 74.00\\
\bottomrule
\end{tabular}
\label{table:ssl+distil}
\vspace{-0.5em}
\end{table}

\begin{table}[t]
\centering
\caption{The impact of different target labels for DiceHuBERT. }
\vspace{-0.5em}
\addtolength{\tabcolsep}{-3.5pt}
\begin{tabular}{lcccc}
\toprule
                          & PR    & ASR (w/o LM) & ASV  & SID \\
\multirow{-2}{*}{Labels} & PER ↓ & WER ↓ / UER ↓        & EER ↓ & ACC ↑\\ \hline\hline 
Hard Labels & 7.48 & \textbf{7.97} / 2.35 & \textbf{5.03} & \textbf{76.85}\\
Soft Labels ($\tau\!=\!1$) & 7.19 & 8.01 / \textbf{2.31} & 5.25 & 76.77\\
Soft Labels ($\tau\!=\!5$) & \textbf{6.70} & 8.02 / 2.33 & 5.51 & 76.36 \\
Soft Labels ($\tau\!=\!10$) & 7.43 &  8.58 / 2.51 & 5.24 & 75.39\\
\bottomrule
\end{tabular}
\label{table:label}
\vspace{-0.4cm}
\end{table}

\vspace{.5em}\noindent\textbf{Distillation Methods.}
We compare our proposed SSL-based distillation method with feature distillation discussed in \cref{sec:feat}. To isolate the impact of architecture, we fix the model architecture to HuBERT-narrow ($D_\text{base}/2$) and apply different distillation methods. For feature distillation loss, we use $\mathcal{L}_{\text{feat}}$ defined in \cref{eq:distil}. We employ the masked input for both $\mathcal{L}_{\text{SSL}}$ and $\mathcal{L}_{\text{feat}}$ to eliminate the influence of the input. The masked input approach also enables the integration of both losses. As evidenced by MaskHuBERT \cite{jang2023recycle}, masking the input does not negatively impact the performance of features distillation.
The results are presented in \Cref{table:ssl+distil}. We find that model trained with SSL loss consistently outperforms the one with feature distillation across all tasks, highlighting the advantages of SSL for distillation. To explore the potential of combining both approaches, we investigate the effect of adding $\mathcal{L}_{\text{SSL}}$ and $\mathcal{L}_{\text{feat}}$ for a training objective, varying the weight for $\mathcal{L}_{\text{feat}}$. Combining the losses yields improvements in SID when the weight is 0.1; however, all other task performance were degraded compared to SSL alone, showing that adding feature distillation atop SSL is not always beneficial.

\vspace{.5em}\noindent\textbf{Target Labels.} In \Cref{table:label}, we investigate the impact of the soft labels introduced for the SSL loss in \Cref{sec:label}.  As HuBERT representations do not lie on the unit sphere, we conduct the experiments by increasing $\tau$ in the soft labels definition to alleviate unnormalized feature space (\cref{eq:softlabel}) and provide reliable targets for a student. We keep the temperature for the student logits as 1, standard default for soft labels distillation~\cite{hinton2015distilling}.
We first see that when the temperature is low ($\tau \!=\!1$), the soft labels are close to hard labels, resulting in similar performance to the model trained with hard labels. The PR shows a slight improvement, but all other tasks are degraded. As the temperature increases, some tasks show improvement (e.g., PR at $\tau\!=\!5$), but the overall performance generally worsens. Notably, when $\tau\!=\!10$, the performance of all tasks was severely degraded, except for a slight improvement in PR. This suggests that over-smoothing the labels is not beneficial for knowledge distillation.

\vspace{.5em}\noindent\textbf{Different Teachers.} To assess the influence of teacher models on our distillation framework, we compare the distillation results obtained using various teacher models in \Cref{table:teacher}.  We employ the first iteration HuBERT-base ($N\!=\!1$), the second iteration HuBERT-base ($N\!=\!2$), and the HuBERT-large model as the teacher models. We pre-trained HuBERT-base ($N\!=\!1$) from scratch, but used pre-trained weights from the official HuBERT repository\footnote{\url{https://github.com/facebookresearch/fairseq}} for the others.  For target generation using $k$-means, we select the 6th layer for HuBERT-base ($N\!=\!1$) and the 9th layer for HuBERT-base ($N\!=\!2$), as used in the HuBERT paper for target generation \cite{hsu2021hubert}. For HuBERT-large, the 18th layer out of 24 was used. 
When HuBERT-base ($N\!=\!1$) serves as a teacher, surprisingly, the student model outperforms all tasks. Despite the teacher model’s poor performance, it generates high-quality targets for the student model that outperform the teacher model even with a smaller model size. These experiments demonstrate the advantages of self-distillation in HuBERT. Furthermore, comparing student performance between HuBERT-base ($N\!=\!1$) and HuBERT-base ($N\!=\!2$) reveals further performance improvement as the teacher’s quality improves. Lastly, when HuBERT-large is used as a teacher, ASR performance improves compared to HuBERT-base ($N\!=\!2$), but not in other tasks. This difference might be attributed to the SUPERB model utilizing all 24 layers of HuBERT-large for downstream tasks, which performs well as a teacher. However, during distillation, only a single layer from the teacher model is selected for target generation, potentially leading to knowledge loss for the student. Future work could explore how to define aggregation between different layers' representations to cluster and provide targets for distillation, e.g. recent work~\cite{chen2024exploring} showed that average of representations instead of selected one layer representation improves HuBERT framework.

\begin{table}[t]
\caption{Performance comparison across different teachers. Note that since HuBERT-large is trained on Libri-Light 60K hours \cite{librilight}, direct comparisons with other models are not entirely fair. However, comparisons can still provide insights into the significance of the teacher model in HuBERT distillation.}
\vspace{-0.5em}
\resizebox{1.0\linewidth}{!}{
\addtolength{\tabcolsep}{-2pt}
\centering
\begin{tabular}{lccccc}
\toprule
                         & \#Param & PR    & ASR (w/o LM) & ASV & SID \\
\multirow{-2}{*}{Models} & M ↓ & PER ↓ & WER ↓ / UER ↓ & EER ↓ & ACC ↑\\ \hline\hline 
\multicolumn{6}{c}{\textit{HuBERT-base ($N\!=\!1$})}\\
Teacher & 95 & 8.56 & 8.92 / 2.62 & 5.58 & 74.33\\
Student & 26 & 7.48 & 7.97 / 2.35 & 5.03 & 76.85\\
\midrule
\multicolumn{6}{c}{\textit{HuBERT-base ($N\!=\!2$})}\\
Teacher$^\dagger$ & 95 & 5.41 & 6.42 / -\hspace{1.5em} & 5.11 & 81.42 \\
Student & 26 & 6.23 & 7.60 / 2.21 & 5.25 & 79.23 \\ 
\midrule
\multicolumn{6}{c}{\textit{HuBERT-large}}\\
Teacher$^\dagger$ & 317 & 3.54 & 3.62 / -\hspace{1.5em} & 5.98 & 90.33 \\
Student & 26 & 6.35 & 6.90 / 2.02 & 5.66 & 76.39\\

\bottomrule
\end{tabular}
}
{\footnotesize $^\dagger$Numbers from the SUPERB paper \cite{yang2021superb}.}
\label{table:teacher}
\vspace{-0.6cm}
\end{table}

 \label{sec:exp}
\section{Conclusion}
In this paper, we introduce DiceHuBERT, a model compression framework that leverages HuBERT's self-distillation mechanism by directly replacing the original model with a student model. DiceHuBERT outperforms feature distillation methods across various SUPERB downstream tasks, achieving over 21\% improvement in ASR and 14\% in PR compared to previous SOTA HuBERT distillation approaches. Additionally, our ablation studies provide valuable insights into designing efficient architectures and selecting optimal distillation targets for future HuBERT compression research.
 \label{sec:conclusion}

\bibliographystyle{IEEEtran}
\bibliography{mybib}

\end{document}